\documentclass[conference]{IEEEtran}
\IEEEoverridecommandlockouts
\usepackage{cite}
\usepackage{amsmath,amssymb,amsfonts}
\usepackage{algorithmic}
\usepackage{graphicx}
\usepackage{textcomp}
\usepackage{xcolor}
\usepackage{float}

\usepackage{xintexpr}
\usepackage{subcaption}
\usepackage{wrapfig}

\def\BibTeX{{\rm B\kern-.05em{\sc i\kern-.025em b}\kern-.08em
    T\kern-.1667em\lower.7ex\hbox{E}\kern-.125emX}}
\begin{document}

\title{Machine Learning Models for Improved Tracking from Range-Doppler Map Images\\
\thanks{Supported by the Defense Advanced Research Projects Agency (DARPA) under contract number HR00112290111. Any opinions, findings and conclusions expressed in this material are those of the authors and do not necessarily reflect the views of DARPA.}
}

\author{\IEEEauthorblockN{1\textsuperscript{st} Elizabeth Hou}
\IEEEauthorblockA{\textit{Systems \& Technology Research (STR)}\\
Arlington, VA 22203, USA\\
elizabeth.hou@str.us}
\and
\IEEEauthorblockN{2\textsuperscript{nd} Ross Greenwood}
\IEEEauthorblockA{\textit{Systems \& Technology Research (STR)}\\
Woburn, MA 01801, USA\\
ross.greenwood@str.us}
\and
\IEEEauthorblockN{3\textsuperscript{rd} Piyush Kumar}
\IEEEauthorblockA{\textit{Systems \& Technology Research (STR)}\\
Woburn, MA 01801, USA\\
piyush.kumar@str.us}
}

\maketitle

\begin{abstract}
Statistical tracking filters depend on accurate target measurements and uncertainty estimates for good tracking performance. In this work, we propose novel machine learning models for target detection \emph{and} uncertainty estimation in range-Doppler map (RDM) images for Ground Moving Target Indicator (GMTI) radars. We show that by using the outputs of these models, we can significantly improve the performance of a multiple hypothesis tracker for complex multi-target air-to-ground tracking scenarios.  
\end{abstract}


\section{Introduction}

Machine learning (ML) systems provide improved accuracy in many data-rich domains like object detection. However, these systems often fail to accurately propagate uncertainty to their outputs given a faithful characterization of the typical input noise -- a feat which traditional statistics-based models in these domains may accomplish straightforwardly. In applications where accurate representation of uncertainties in outputs is crucial e.g., a statistics-based estimator such as a Kalman filter, the superior accuracy of such ML systems is not useful due to the lack of reliable estimates of their outputs' variability.

In this paper, we develop a novel detection system consisting of coupled machine learning models: i) a UNet based neural network architecture that gives improved target detection performance (w.r.t. CFAR baseline) in simulated air-to-ground scenarios, and ii) a conditional variational autoencoder (CVAE) based neural network that estimates the uncertainties of the UNet model's predictions. Rather than propagate input uncertainty through the UNet model layers or evaluate UNet multiple times for each input to build sample statistics, the CVAE learns the output distribution conditioned on the UNet's penultimate layer features. This allows us to predict the output distribution using a single new sample (from an intrinsic input distribution), whereas uncertainty propagation and direct sampling do not\footnote{In the latter cases, many samples are needed to estimate input distribution for propagation or feed through the UNet to generate output samples.}. We show that by replacing traditional methods with our more powerful machine learning models in a detection and tracking pipeline, we are able to significantly improve the performance of the tracker in complex multi-target tracking scenarios.

\subsection{Related Work}

Detecting targets in range-Doppler map (RDM) images from airborne radars is a relatively ``niche” domain limited to mostly defense applications. Unlike more mainstream domains such as electro-optical (EO) or even infrared (IR) where there are prolific amounts of literature (from the machine learning community) on object detection in RGB images, there is very little published literature on machine learning models for detection in RDM images. The authors in \cite{roldan2019dopplernet} and \cite{kim} both use convolutional neural networks (CNNs) to classify humans from other types of objects (e.g. cars, dogs, drones) in RDM images. And, the authors in \cite{peng} and \cite{kim2} also use CNNs to classify human activities from micro-Doppler and range-Doppler signatures. However, even in these other works, they are focused on ground radars with short-range distances as opposed to airborne radars taking ``overhead" imagery at long-ranges. 

Uncertainty estimation in neural networks is a bit more well studied; however, uncertainty estimation models will naturally depend on the specific neural network whose output distribution its estimating. \cite{Gawlikowski} is a survey paper that gives a comprehensive overview of uncertainty estimation in neural networks. Most similar to our proposed model is \cite{cvae} who also propose a CVAE architecture, but for the purposes of modeling uncertainty in image restoration. 

\subsection{Outline}
In the following sections, we first describe the tracking scenario in Section \ref{scenario}. Then we 
illustrate the neural network architectures for our target detector and uncertainty estimation models in Sections \ref{unet} and \ref{cvae} respectively. Since these models do not have a time component, we drop the $t$ subscripts in these sections for clarity and simpler notation. Finally in Section \ref{experiments}, we first describe our method for simulating training and test datasets, and then show performance results on these datasets. 

\section{Problem Definition} \label{scenario}

We are interested in a scenario in which a radar on an airborne platform is collecting measurements of targets on the ground. The airborne platform's position is known in Cartesian coordinates affixed to the ground, i.e. East, North, Up (ENU). The \textit{latent} state $z$ of each target is taken to be a 6 dimensional vector that encompasses its position and velocity in ENU coordinates. It is modeled as evolving linearly over time according to some dynamics matrix $\Phi$ and with some additive white noise  $\omega \sim N(0, Q)$. Each target's measurements $y$ are a 4 dimensional vector consisting of its range, range-rate or radial velocity (Doppler), azimuth, and elevation angle relative to the airborne platform's position and a reference orientation vector directed broadside. We can transform from the targets' latent states $z$ to their measurements $y$ with a function $h(\cdot)$ or a linear approximation $H$ (see Appendix Section \ref{h_func} for more details) and some additive white noise $\epsilon \sim N(0, R_t)$ representing the measurement uncertainty due to the inherent noise from a sensor, e.g. thermal noise from machinery.

For each target, we can model the system as the following
\begin{flalign}
& \text{Dynamics Model: } z_t = \Phi z_{t-1} + \omega_t \label{dynamics} \\
& \text{Measurement Model: } y_t = H_t z_t + \epsilon_t \label{measurement}
\end{flalign}
where $y_t$ are typically \textit{observable} measurements and  $\Phi, H_t, \omega_t$ and $\epsilon_t$ are typically known or given. However, a radar or other external sensor cannot directly measure a target's range, range-rate (Doppler), azimuth, or elevation. Instead, it can take measurements of an area containing the targets, which can be processed into images with targets within them. Specifically in this paper, we are considering Ground Moving Target Indication (GMTI) radars whose raw measurements are processed into RDM images. Thus, we propose a novel neural network model that detects targets $\hat{y}_t$ from observable measurements $X_t$ (RDM images) to use as estimates of $y_t$ in the measurement model above. Also while we may be able to quantify the noise in the sensor taking the measurements $X_t$, we do not know its distribution after a complex transformation into ``$y$ space". Thus, we also propose to learn this transformation with another neural network such that given a distribution of images $X_t$ it learns the transformed distribution of its noise in range and range-rate. This statistical model of the target detector is can be intuitively though of as approximately learning the distribution of $\epsilon_t = f(e_t)$ where $f$ is the function approximated by the target detector neural network $\hat{y}_t = f(X_t)$ and $e_t$ is the sensor noise in the RDM images.

\section{Target Detection Model} \label{unet}

\subsection{Challenges}

Target detection in RDM images has some unique challenges not present in traditional object detection problems. The targets in these images are approximately only a single pixel in size and don’t have structural components (e.g. edges, object features) to detect. Traditional methods for detection in this domain are essentially statistical threshold tests of whether each pixel contains a target or not, which shares more similarities with image segmentation than typical object detection. Consequently, we adapted the UNet architecture from image segmentation models (originally proposed in \cite{unet}) as the backbone of our model's architecture.

\begin{figure}[H]
    \centering
    \includegraphics[width=0.75\linewidth]{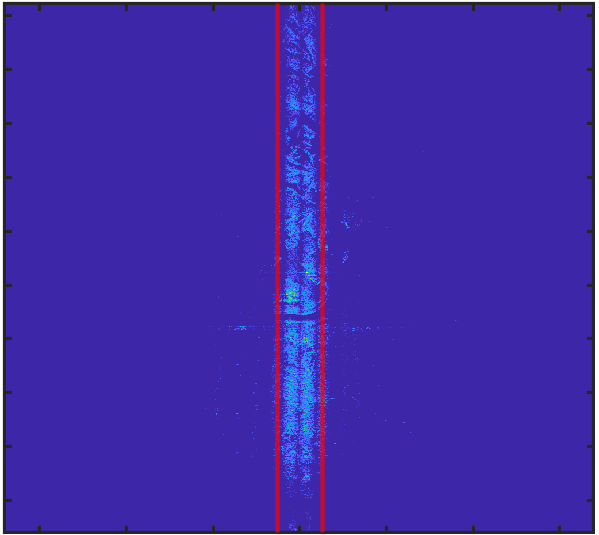}
    \caption{The red lines indicate the endo-exo divide separating the endo-clutter region inside the lines with the exo-clutter region outside the lines.} 
    \label{fig:clutter}
\end{figure}

Another challenge, specific to airborne GMTI radars, is that the measurements will include ground clutter in addition to additive white noise from the sensor. For RDM images from monostatic radars, this comes in the form of significant additional noise in the center of the images, an example is shown in Figure \ref{fig:clutter}.
In order to remove this noise from ground clutter, space time adaptive processing (STAP) is typically used to whiten it out, but this can be computationally expensive and increasingly difficult when the radar has numerous channels. Thus we leverage the power of discriminative methods and use supervisory signals to train our model to learn to detect targets even in sophisticated noise regimes. 

\subsection{UNet Architecture}

The target detector model takes in labelled training data $\{X_n, Y_n\}$ in the form of a complex valued RDM image $X_n$, which is a $h \times w \times m$ complex matrix where $h$ and $w$ are the pixel dimensions and $m$ is the number of channels in the radar, and a corresponding $h \times w$ binary matrix $Y_n$ indicating whether each pixel contains a target or not. 

\begin{figure}[H]
    \centering
        \includegraphics[width=0.85\linewidth]{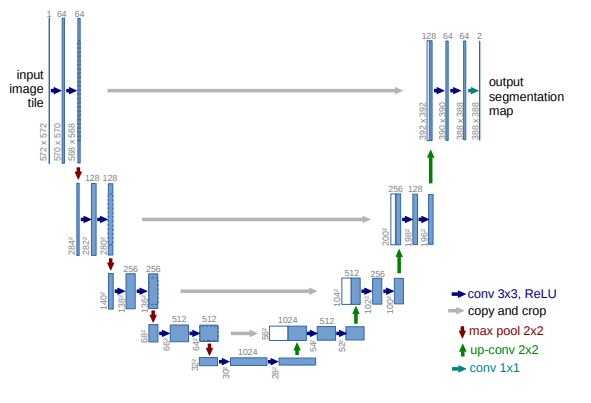}
    \caption{Example of a UNet architecture: Blue boxes are multi-channel feature maps, white boxes are copied feature maps, and the arrows denote the different operations.} 
    \label{fig:unet}
\end{figure}

The backbone of the detector has a UNet architecture (a generic version from \cite{unet} is shown in Figure \ref{fig:unet}) consisting of 6 blocks in the contraction path and 6 blocks in the expansion path. Each block consists of two layers of convolution operations with a ``depth" of $32$, followed by a pooling or up-sampling operation with concatenation, and finally a dropout layer. The goal of the contraction path is the ``standard" convolutional goal of pooling information from pixels together to learn feature maps. The goal of the expansion path is to use these learned features maps and concatenate them together with pixel information to ``expand" or up-sample the features maps to the pixel resolution. Thus the output of the UNet backbone (the penultimate layer of the full model architecture) is a $h \times w \times 32$ tensor and can be interpreted as each pixel having $32$ unique learned features. While each pixel has its own feature vector, this vector will include information from neighboring pixels due to the convolutions layers within the blocks. 
The final layer or ``task head" of the detector is a convolution with kernel size 1 and a soft-max activation function. This is essentially a feed-forward layer (with the same shared weights) applied independently to each pixel's $32$ dimensional feature vector from the backbone. The target detector model is trained so that the outputs of the final layer minimize a weighted cross entropy loss function $\mathcal{L}(Y, \hat{Y})=$
\begin{flalign} \label{unet_loss}
-\sum_{i,j} \omega^1 Y_{(i,j)} \log \hat{Y}_{(i,j)} + \omega^0 (1 - Y_{(i,j)}) (1 - \log \hat{Y}_{(i,j)})
\end{flalign}
where $\hat{Y}_n$ is a $h \times w$ matrix of the outputs of final layer and the weights $\omega^0 = \frac{h*w}{(h*w) - K}$ and $\omega^1 = \frac{h*w}{K}$ are the average number of targets $K$ in each image in the training set. This is necessary due to the large class imbalance as the number of pixels without targets (0's) can be much larger than the number of pixels with targets (1's). And without balancing the loss function, the target detector would degenerate and learn to just predict everything as the 0 class. 

\begin{figure}[H]
    \centering
        \includegraphics[width=0.8\linewidth]{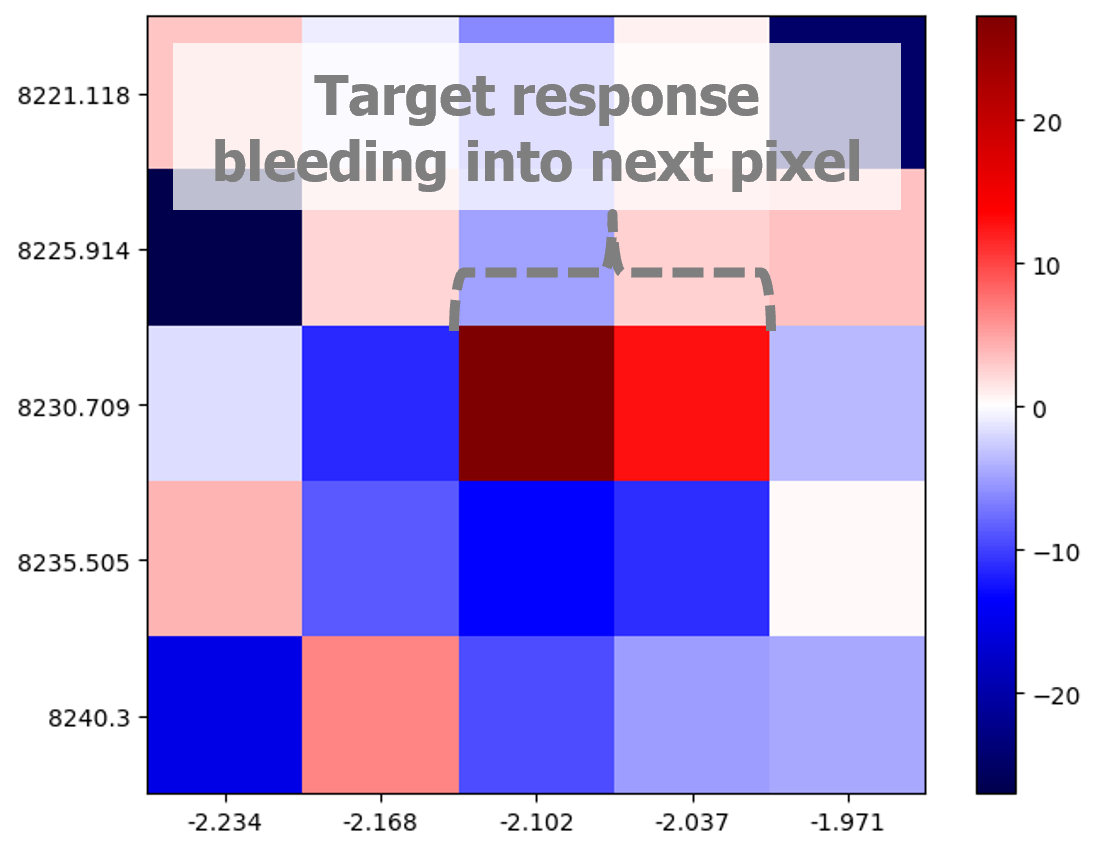}
    \caption{Example of the target response (dark red center pixel) in a RDM image bleeding into its neighboring pixel (bright red) due to the discretization from binning of the target response.} 
    \label{fig:blur} 
\end{figure}
While $\hat{Y}_n$ is useful for measuring the accuracy of our target detector, we are interested in detecting targets to use as estimators for the $y$'s in the measurement model \eqref{measurement}, which are in range and range-rate. Let $\{ \hat{y}^k \}_{k=1}^K$ be a list of $K$ targets where each $\hat{y}^k$ is a $2 \times 1 $ vector with a range and range-rate element and corresponds to one of predicted $1's$ in the threshold-ed $\hat{Y}$ matrix. We can simply assign each $\hat{y}^k$ the associated range and range-rate bins of the pixel. However, because RDM images are capturing aspects of a ``continuous" real world in discrete sensor measurements, targets may not fall exactly within a pixel bin and instead between pixels. For example, in Figure \ref{fig:blur}, we illustrate this where a target response in the center-most pixel is bleeding into a neighboring pixel on the right. 

Thus in order to more accurately estimate the range and range-rate of a target, we use a weighted averaging method. Explicitly
\begin{flalign} \label{weighted_avg}
\hat{y}^k = \frac{\sum_{(i,j) \in W(k)} s_{(i,j)} \hat{Y}_{(i,j)}}{\sum_{(i,j) \in W(k)} \hat{Y}_{(i,j)}} 
\end{flalign}
where $W(k)$ is a patch of pixels centered around predicted target $k$, $s_{(i,j)}$ are the range and range-rate coordinates corresponding to pixels in the patch, and $\hat{Y}_{(i,j)}$ are the post soft-max probabilities used as weights for each coordinate. For example, using the target shown in Figure \ref{fig:blur}, the predicted target location using the corresponding coordinates of the center pixel is $(8230.709, -2.102)$; however, using \eqref{weighted_avg} its $(8230.508, -2.091)$. The latter method adjusts for the target response's discretization into multiple pixels. 

\section{Statistical Model of the Target Detector} \label{cvae}

A common method for modelling probability distributions with neural networks is using a variational autoencoder architecture, which is trained by learning weights that maximize the likelihood of the data fitting the distribution. However, because we are interested in estimating the distribution of the outputs of the target detector model \textit{given} a specific RDM image, we are interested in conditional distributions. So we use a variant of the architecture called a conditional variational autoencoder (CVAE), which will learn weights that maximize the likelihood of the conditional distribution of the transformed noise of a RDM image. Modeling residuals distributions with a CVAE for estimating uncertainty was first proposed in \cite{cvae} for the purpose of image restoration. We build off their framework and design a deep neural network with a similar architecture. 

Specifically, our model takes in the same labelled training data $\{X_n, Y_n\}$ as the UNet target detector. It then embeds the $h \times w \times m$ complex RDM image matrix into the pre-trained, fixed features maps of the penultimate layer of the UNet target detector. For training, the model only wants to train on correct information, thus we only use the residuals of the correctly predicted targets $r_{(i,j)} = y_{(i,j)} - \hat{y}_{(i,j)}$ and the corresponding $32 \times 1$ feature vectors $h_{12}(X_{(i,j)})$ of those $(i,j)$ pixels where $h_{12}(\cdot)$ is the pre-trained UNet backbone. We assume that these residuals $r$ are instantiations of the sensor noise $\epsilon$ in \eqref{measurement}.

The model minimizes the variational lower bound to the conditional distribution of residuals given images for the model's loss function $\min_{\psi, \phi, \Sigma} \mathcal{L} =$
\begin{flalign} \label{cave_loss}
KL(q_\phi(\rho| X, r) || p_\psi(\rho|X)) - E_{\rho\sim q_\phi}(\log P_\Sigma(r|X, \rho)) 
\end{flalign}
where $\psi, \phi, \Sigma$ are the parameters of the distributions modelled by the encoder, reference, and decoder blocks respectively. The function $KL(\cdot || \cdot)$ is the Kullback Leibler (KL) divergence and the expectation $E_{\rho \sim q_\phi}$ is the expectation on the latent variable $\rho$ with respect to the reference distribution $q_\phi(\rho| X, r)$. The goal of the encoder block is to learn a Gaussian distribution $p_\psi(\rho | X)$ for a lower dimensional latent variable $\rho$ using only the penultimate layer feature vectors $h_{12}(X)$. The goal of the reference block $q_\phi(\rho| X, r)$ is to similarly learn a Gaussian distribution for $\rho$, but with additional information in the form of training labels $r = y - \hat{y}$ (therefore it is useful for training, but not prediction). Finally, the goal of the decoder block is to model $P(r|X, \rho)$, a Gaussian distribution for the residuals $r$ conditioned on the latent $\rho$ and the penultimate layer feature vectors. Each block consists of a series of hidden layers where each of these has a feed-forward layer followed by normalization and a ReLu activation function.

\begin{figure}[H] 
    \centering
    \includegraphics[width=0.98\linewidth]{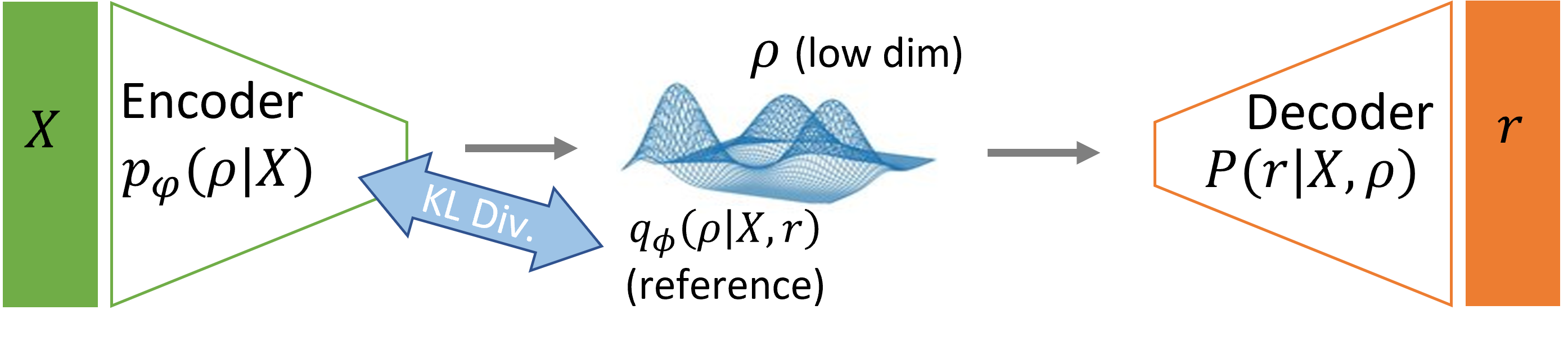}
    \caption{The architecture of one of the twin (endo- and exo-clutter) CVAEs in the statistical model of the target detector.} 
    \label{fig:cvae}
\end{figure}

However, the pixels in the endo-clutter region also have noise from the ground clutter returns. Thus we want to estimate different distributions for pixels in the endo and exo-clutter regions. We can duplicate the CVAE architecture so that there are 6 blocks in total i.e. an encoder, reference, and decoder block for modelling both the endo-clutter distribution and the exo-clutter distribution. Figure \ref{fig:cvae} shows the architecture of one of the CVAEs
\noindent where the model's total loss function is just the sum of two copies of \eqref{cave_loss}. Note that given metadata from monostatic radars, e.g. the direction the radar is pointing, we can roughly estimate which pixels lie in the endo-clutter region. 

The goal of the loss function in \eqref{cave_loss} is to learn good estimators of the encoder $p_\psi(\rho | X') \partial$ and decoder $P_\Sigma(r' | X', \rho)$ distributions. It enforces this by ensuring that the encoder distribution is close in KL divergence to the reference distribution (first term) and the parameters of the decoder distribution minimize its expected (with respect to the reference distribution) negative log likelihood (second term). Then a prediction time, for both detections in the endo and exo-clutter regions, the corresponding trained encoder and decoder blocks are convolved such that given a new test image $X'$, 
\begin{flalign} \label{r_dist}
    P(r' | X') = \int P_\Sigma(r' | X', \rho) p_\psi(\rho | X') \partial \rho
\end{flalign}
where $P(r' | X')$ is the transformed uncertainty given a specific input $X'$. From this mixture model decomposition, we can see that intuitively the lower dimensional latent variable $\rho$ is further encoding information necessary for representing the distribution of the residuals $r'$. We can efficiently generate realizations of $r'$ by sampling from the encoder and decoder distributions. The sample covariance of these realizations is an estimator $\hat{R}$ for the measurement uncertainty $R$ in \eqref{measurement}; the number of samples will dictate the ``goodness-of-fit" of this covariance estimator according to the convergence rates of sub-Gaussian distributions \cite{vershinyn_paper}. Additionally, $P(r'|X')$ is in general more expressive than a single Gaussian parameterized by $\hat R$, and could likely be used more efficiently by more sophisticated estimators; however higher order moment estimators would also require more samples for a good estimation. In general though, sampling is very computationally efficient as each sample is a parallelizable forward pass through the neural networks of the encoder and decoder blocks.  

\section{Experiments} \label{experiments}

\subsection{Simulated Data Description}

In order to train our proposed neural networks, we need large amounts of RDM image, target location pairs $\{X_n, Y_n\}$ for labelled training data. Since large real datasets of RDM images with labelled target locations do not exist, we built a sensor model for our experiments to generate such datasets. We assume each RDM is the $h \times w \times m $ realization of a complex random variable $x$ that can be decomposed as
\begin{flalign} \label{eq:sensor_model}
x = x^{tgt} + x^{clutter} + x^{noise}
\end{flalign}
where the last two components are parameterized by environmental and radar parameters such as the sensor's waveform and thermal noise and the clutter due to reflectivity of the ground. Since these environmental factors are inherently captured in sensor measurements, we use a few samples of real RDM images (no target location pairings) to estimate these components in our sensor model. This allows us to generate additional simulated RDM images with target location pairs that are sufficiently realistic. 

We assume that the clutter and noise random variables come from zero mean complex Gaussian distributions where the noise covariance for each pixel component $Cov(x^{noise}_{(i,j)})$ is a diagonal matrix (channels are independent), and the clutter covariance for each pixel component $Cov(x^{clutter}_{(i,j)})$ is a dense $m \times m$ matrix (channels are not independent). We can estimate these covariances by first dividing the pixels in a real RDM into those that only contain the noise component and those that contain both clutter and noise components. Since the noise random variables are identically distributed for all $i, j$ pixels, i.e. $Cov(x^{noise}_{(i,j)}) = \Sigma^{noise}$, we can use all pixels in the exo-clutter region of a real image as samples to estimate its covariance. For each channel, we take the median of $X^o_{i,j} X^{o\prime}_{i,j}$ (where $'$ is the complex conjugate) and perform the Exponential distribution median to mean conversion to get a sample variance estimate. The clutter random variables have a different covariance $Cov(x^{clutter}_{(i,j)}) = \Sigma^{clutter}_{(i,j)}$  for each pixel and we only have one realization of each pixel in a real image. So we assume the distributions across the pixels is slowly varying over the range axis and use a moving window to form samples for estimation. Then the estimator for the covariance matrix of $\Sigma^{clutter}_{(i,j)}$ is just the sample covariance of all the pixels in its moving window. Note, we only need to estimate these moving window sample covariances for pixels in the endo-clutter region; for pixels in the exo-clutter region, we assume the clutter covariance is degenerate (zero).

The sensor model injects targets into the simulated RDM images using the radar parameters corresponding with a real image and physics-based equations. For training our neural networks, we just need large numbers of independent labelled images to form a training dataset. So we can just randomly choose the locations of $K$ targets in a generated image where each target is assigned a signal value drawn with respect to a signal-to-noise ratio (SNR) parameter, transformed with physics equations, and adjusted by its angle of intersection with the radar beam. For our experiments below, we chose $K=300$ targets with approximately half of them lying in the endo-clutter region in order to provide our target detector with enough examples of positive detections (1's class) in more complex regions of the image. However, for evaluating the effectiveness of our proposed neural networks in improving target tracking, we need a time series of target, image pairs. 

We simulate targets that move over time by embedding a scenario within the sensor model. For a simple scenario, we use a constant velocity model where we set the dynamics matrix $\Phi$ in \eqref{dynamics} so that both the ground targets and the airborne platform move with 
\begin{flalign} 
\Phi = 
\begin{bmatrix}
\mathbf I_3 & dt \, \mathbf I_3 \\
\mathbf 0 & \mathbf I_3
\end{bmatrix}
\end{flalign}
where $dt$ is the time interval and $\mathbf I_3$ is the identity matrix. For a more complex scenario, we can have the airborne platform circle the ground targets in combat air patrol mode and the targets can accelerate, decelerate, stop, and turn. In order to ensure that the targets stay within the radar's field of regard for the entire scenario and are covered by the radar beam every couple of time points, we sweep the radar's aim azimuth with electronic steering. For the tracking experiments below, we chose $K=8$ targets for both scenarios where the targets will move in and out of the endo-clutter region as time elapses. 

Once we simulate the platform and target locations and velocities at each time point in Cartesian space, we calculate the targets' relative range and range-rate value (according to \eqref{range} and \eqref{rangerate}, respectively) and discretize them to their nearest corresponding range and range-rate bins, where the bins correspond directly with pixels in the final RDM image. This provides us with labels $Y_{(i,j)}$ of the "most prominent" pixel location for each target, i.e. a 0-1 matrix of whether each $(i,j)$ pixel contains a target or not, in order to train our UNet base target detector with the weighted cross entropy loss function in \eqref{unet_loss}. However; the signal of the targets' will bleed into neighboring pixels in the RDM images and we use \eqref{weighted_avg} to recover the actual (non-discretized) target range and range-rate values. We also calculate the targets' azimuth with \eqref{azimuth}, which is used to calculate each target's signal value at each time point. With these three components assigned to each target, they are injected into the sensor model in a similar fashion as before. Note that the targets' signal return is dependent on it and the radar's azimuths. Thus, a sweeping radar azimuth and varying measurement window is necessary so that there will be a detectable signal return at some time point within every couple of radar measurements.

\subsection{Target Detector Accuracy}

In order to evaluate the performance of our target detector, we compare its accuracy to the traditional method of pre-processing the image with space time adaptive processing (STAP) and then using a constant false alarm rate (CFAR) test. The CFAR test is a variation of the Neyman-Pearson statistical hypothesis test \cite{cfar1, cfar2} where the null hypothesis is that there is no signal present in the pixel and the alternate hypothesis is that there is a signal. It assumes all pixel intensities are independent and identically distributed (i.i.d.) from a complex Gaussian distribution with zero mean and unknown covariance. In practice, this unknown covariance is estimated with maximum likelihood and the test is a generalized likelihood ratio (GLR) test, which asymptotically converges to a CFAR test \cite{asym}.

For these experiments, we trained our UNet based target detector on 8,000 simulated images generated from 8 real images for 22 epochs on a NVIDIA Quadro RTX 6000 GPU. Then we used 400 simulated images generated from different 4 real images as a test set for evaluating accuracy. Since the traditional method of STAP+CFAR does not require any training, we just applied it to the same test set.

\begin{figure}[H] 
    \centering
        \includegraphics[width=\linewidth]{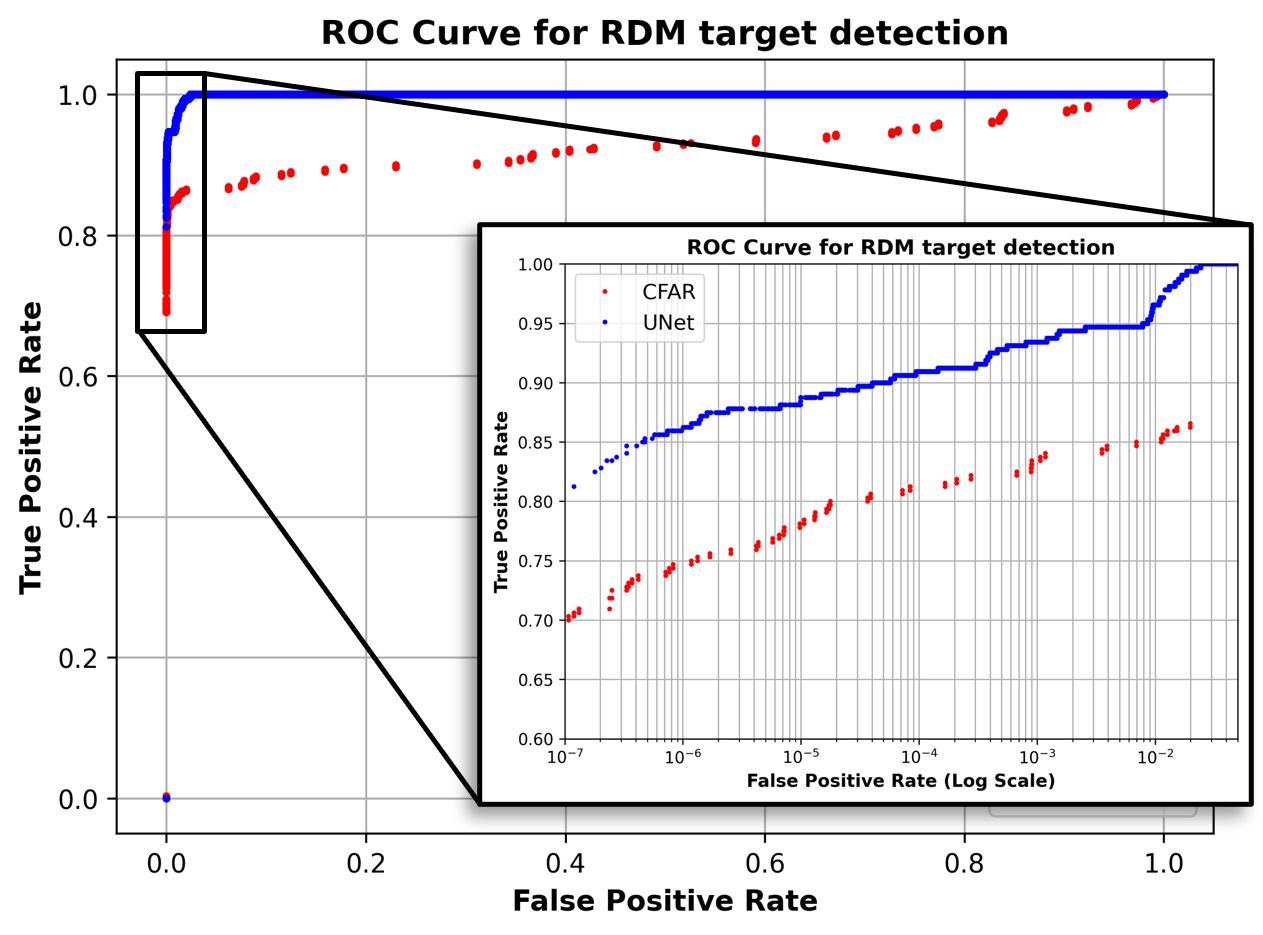}
        \caption{Receiver Operating Characteristic (ROC) curve: Red is the traditional STAP prepossessing followed by the CFAR test, Blue is our UNet based target detector. Inner figure is plotted with the false positive rate (FPR) on a log scale.}
        \label{fig:roc}
\end{figure}

Figure \ref{fig:roc} shows the true positive rate (TPR) as a function of the false positive rate (FPR) for the two methods. At every FPR, the UNet based target detector is equal or more sensitive than the traditional non-machine learning CFAR method. For example, if we look at the performance at a typical $0.05$ alpha level test (i.e. FPR is 0.05), we see that the UNet method has TPR approaching $1.0$ while the power of the CFAR test is only $0.8889$. Or if we look at a much lower level of $10^{-6}$ corresponding to approximately $2.1$ false detections per image, the UNet detector has a TPR of $0.86$ whereas CFAR is lower at $0.76$. In order to highlight the UNet detector's performance in regimes with realistic numbers of false detections (very low FPR), we zoom into this area and plot the FPR on a log scale in the inner box of the figure. 

While the CFAR test is \textit{theoretically} a uniformly most powerful test, this is only true when the distributional assumptions (e.g. pixels are i.i.d. from complex Gaussian) are not violated. Our UNet based target detector on the other hand is a discriminative model that uses labelled data to learn feature vectors for each pixel where these features vectors contain information from neighboring pixels. Thus by not imposing any distributional assumptions, but instead implicitly learning the distribution using the supervisory signal from labelled data, our UNet based target detector is able to be more accurate than the traditional CFAR test.   

\begin{figure}[H]
    \centering 
    \includegraphics[width=\linewidth]{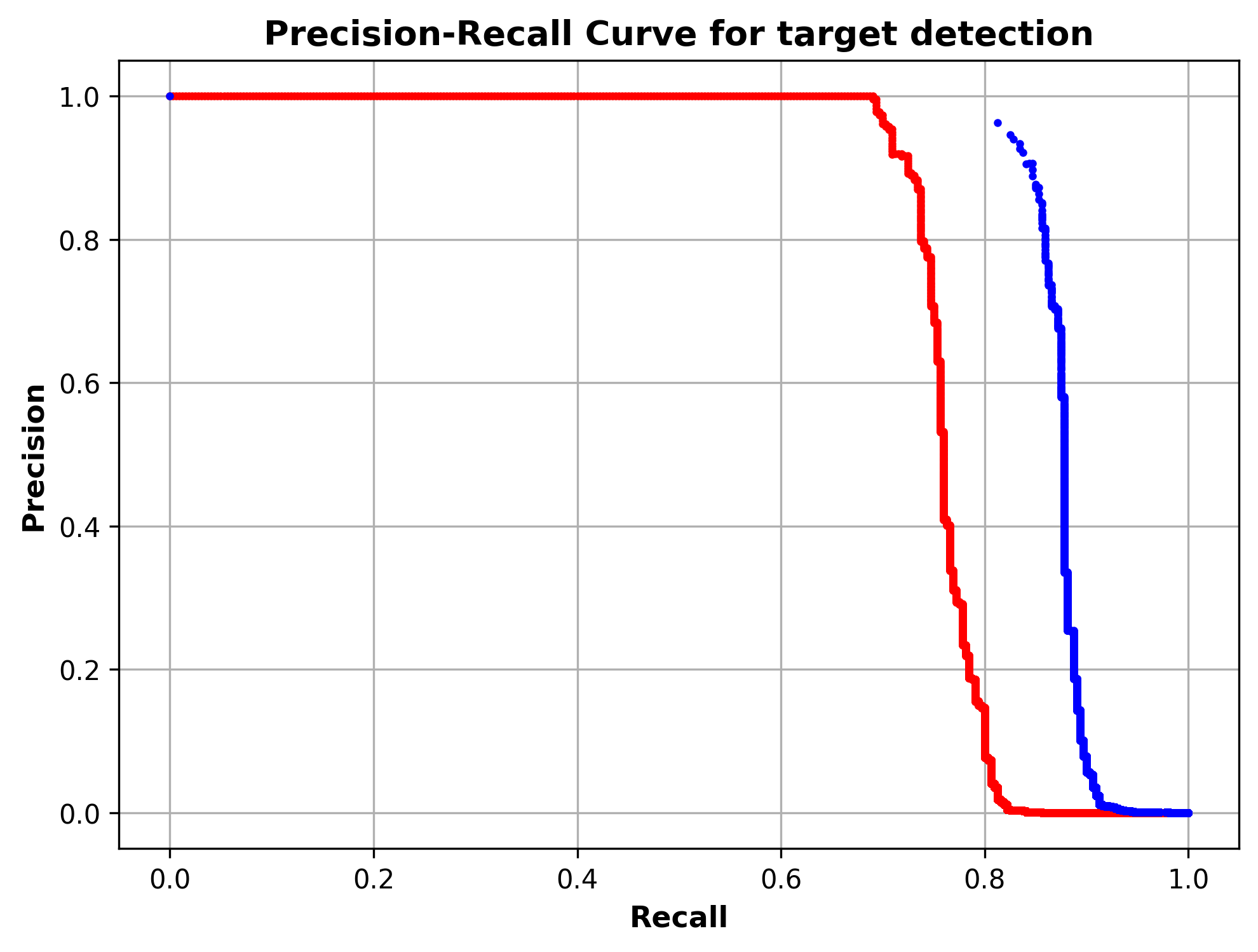}
    \caption{Precision Recall (PR) curve: Red is the traditional STAP prepossessing followed by the CFAR test, Blue is our UNet based target detector. }
    \label{fig:pr}
\end{figure}

\begin{figure}[H] 
    \centering
    \includegraphics[width=0.9\linewidth]{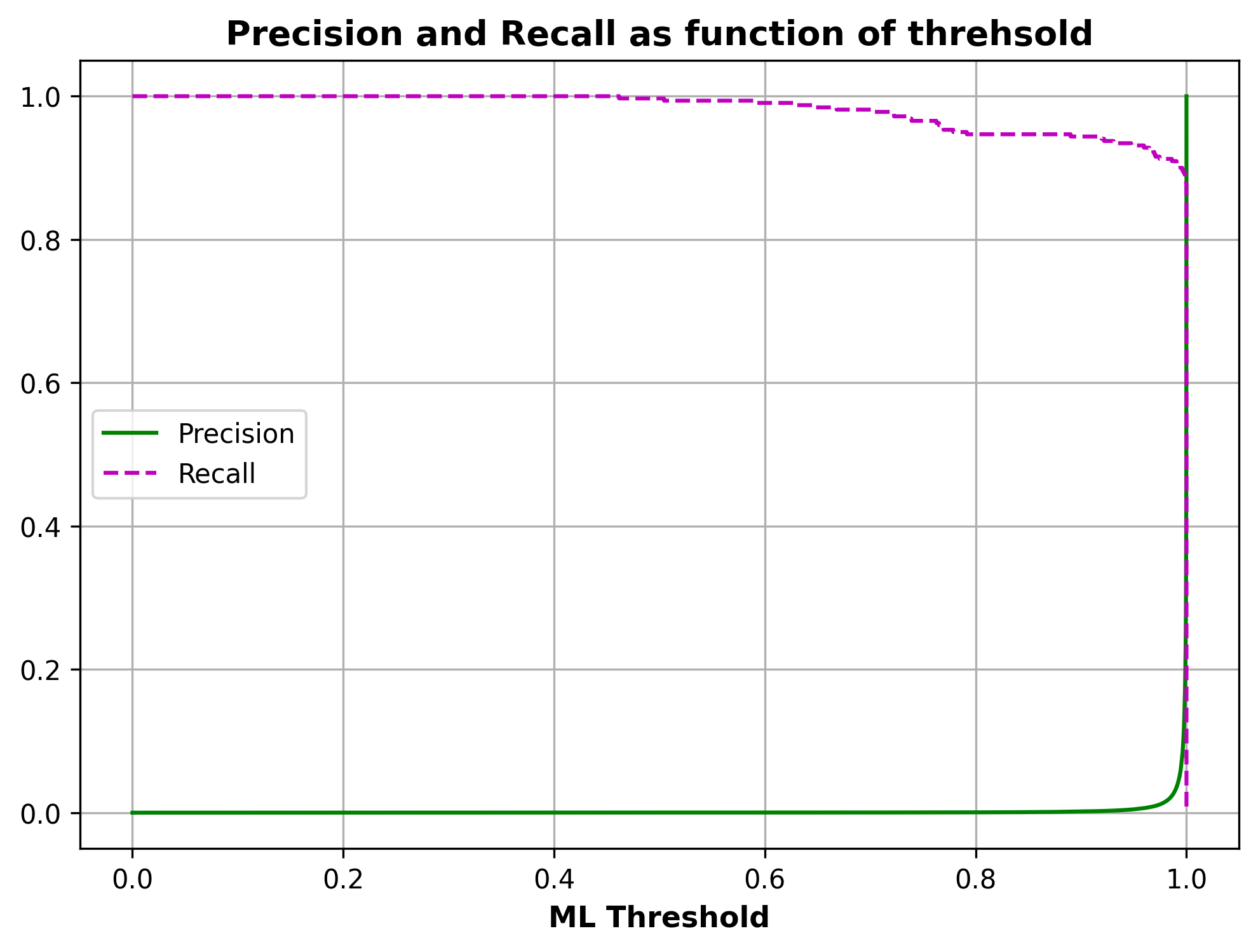}
    \hfill
    \includegraphics[width=0.9\linewidth]{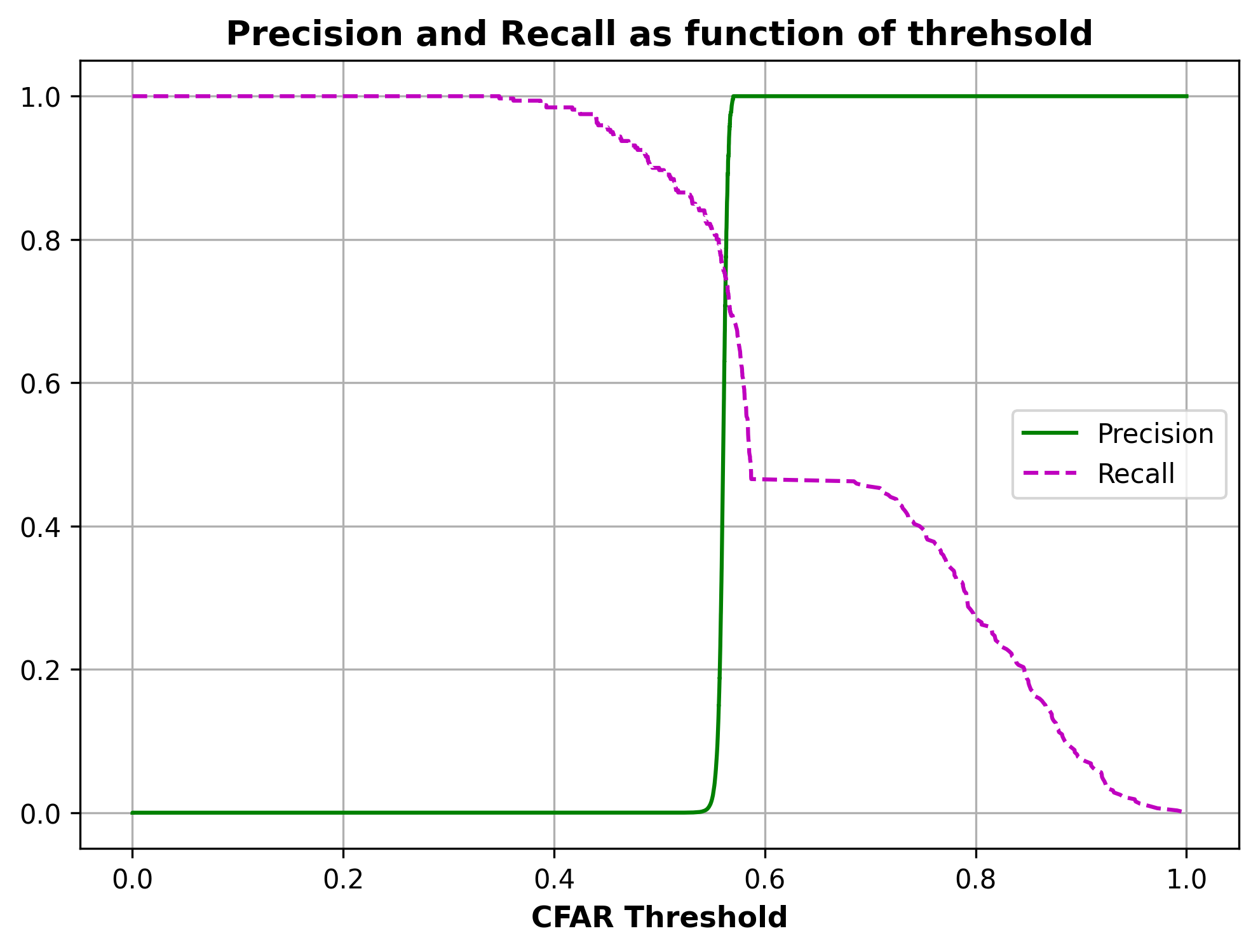}
    \caption{Precision (green) and Recall (pink) as a function of threshold for the UNet target detector (top) and the traditional STAP prepossessing followed by the CFAR test (bottom).}
    \label{fig:thresh}
\end{figure}

Figure \ref{fig:pr} shows the precision as a function of the true positive rate (recall) for the two methods. Because the UNet target detector is not a statistical test, it will not always have a smooth output to threshold at. So while the CFAR test slowly increases its detections as the threshold decreases, the UNet model immediately predicts a large chunk of detections at once and the PR curve ``jump starts" to a TPR of $0.8125$. At this TPR, the UNet target detector is much more precise, at $0.963$, than the competing CFAR test which only has precision $0.0354$. To provide more clarity, we also plot the precision and recall as a function of threshold in Figure \ref{fig:thresh} where we can see that the UNet model has lots of high probability predictions that are detected even at a very high threshold whereas the CFAR test has a much more gradual detection rate as a function of threshold. For example, at a threshold equal to $1.0$ (the ``jump start"), the UNet target detector has a precision of $0.963$ and a recall of $0.8125$. However, in order to have the same level of precision for the CFAR test, its recall is lower at $0.7$. Thus we see similar performance as the ROC curve where the UNet based target detector is more accurate compared with the CFAR test for various levels of precision. 

\subsection{Improving Tracking Accuracy} 

Now that we have shown the superior performance of our target detector individually, we can also evaluate its contribution to improving target tracking. Since a statistical tracker also needs uncertainty estimates, the results below also highlight the necessity and performance of our statistical model of our target detector. We compare the performance between using estimates from our target detector and its statistical model with estimates from traditional methods. Since our experiments involve multiple targets, we use a multiple hypothesis tracker (MHT), which has been well established as a leading paradigm for solving the multi-target tracking problem \cite{mht1,mht2,mht3,mht4,mht5,mht6}. 

The two systems we compare on both the simple and complex scenarios are:
\begin{enumerate}
\item ML-Filter: Trained UNet based neural network for target detection, Trained CVAE based neural network for uncertainty estimation, and MHT for target tracking
\item Baseline-Filter: STAP for pre-processing, Constant False Alarm Rate (CFAR) model for target detection, a constant covariance matrix shown in \eqref{base_cov}, and MHT for target tracking
\end{enumerate}
For CFAR, we applied a threshold of 30 to the post-STAP pixel values; if the normality/whiteness assumptions were satisfied\footnote{If STAP complex pixel values were distributed as white noise,
the scores would be governed by a chi-squared distribution with 2 degrees of freedom.  
}, this would correspond to a FPR of $3.1\times10^{-7}$, or about 0.64 false detections per $2048\times1024$ RDM image on average (well within the multiple hypothesis tracker's filtering capability). The UNet threshold was chosen empirically to give the tracker a comparable number of true detections.

For evaluation, we use the five metrics established in \cite{mht-metrics} that collectively provide a good overall assessment of the performance of predicted tracks in a multi-target setting. Table \ref{table:tracking_results}, shows that our proposed machine learning methods improve the performance of the tracker in both target and track purity metrics ($TaP$ and $TrP$) with significant performance in the case of localization error ($LE$). It has slightly worse in performance in target completeness ($TaC$), but significantly better performance in track completeness ($TrC$) in the the complex scenario. 
\begin{table}[H]
\caption{Performance in Target Tracking}
\begin{center}
\begin{tabular}{|l || c c || c c |} 
 \hline
Metric & \multicolumn{2}{c||}{\underline{Constant Velocity (Simple)}} & \multicolumn{2}{c|}{\underline{MoveStopMove (Complex)}} \\ 
  & ML-Filter & Baseline-Filter & ML-Filter  & Baseline-Filter \\ [0.5ex] 
 \hline\hline
 $TaC$ & 0.5075 & 0.5127 & 0.9482 & 0.9848 \\ 
 \hline
 $TrC$ & 1 & 1 & 1 & 0.6476  \\
 \hline
 $TaP$ & 0.9639 & 0.9428 & 0.9050 & 0.8756  \\
 \hline
 $TrP$ & 1 & 0.995 & 0.9816 & 0.9685 \\
 \hline
 $LE$ & 0.007 & 0.114 & 0.0101 & 0.1403 \\ 
 \hline
\end{tabular}
\end{center}
\label{table:tracking_results}
\end{table}
\noindent where $TaC$ is Target Completeness, $TrC$ is Track Completeness, $TaP$ is Target Purity, $TrP$ is Track Purity, and $LE$ is Localization Error in kilometers. The first four metrics are between $[0, 1]$ and higher is better, while for the last metric, lower is better.

\begin{figure}[H]
    \begin{subfigure}{0.49\linewidth}
    \centering
    \includegraphics[width=\linewidth]{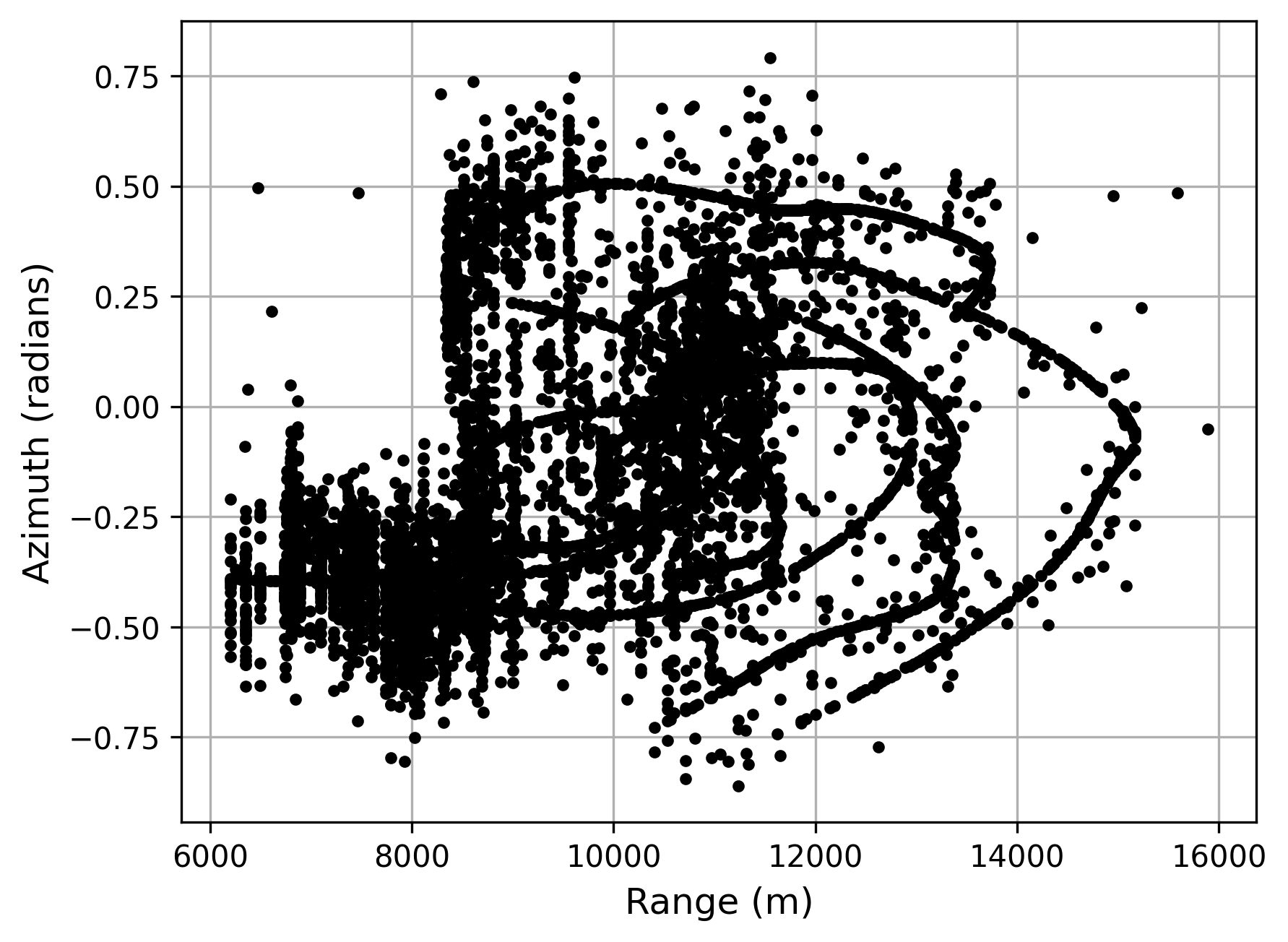}
    \subcaption{UNet Target Detector}
    \end{subfigure}%
    \hfill
    \begin{subfigure}{0.49\linewidth}
    \centering
    \includegraphics[width=\linewidth]{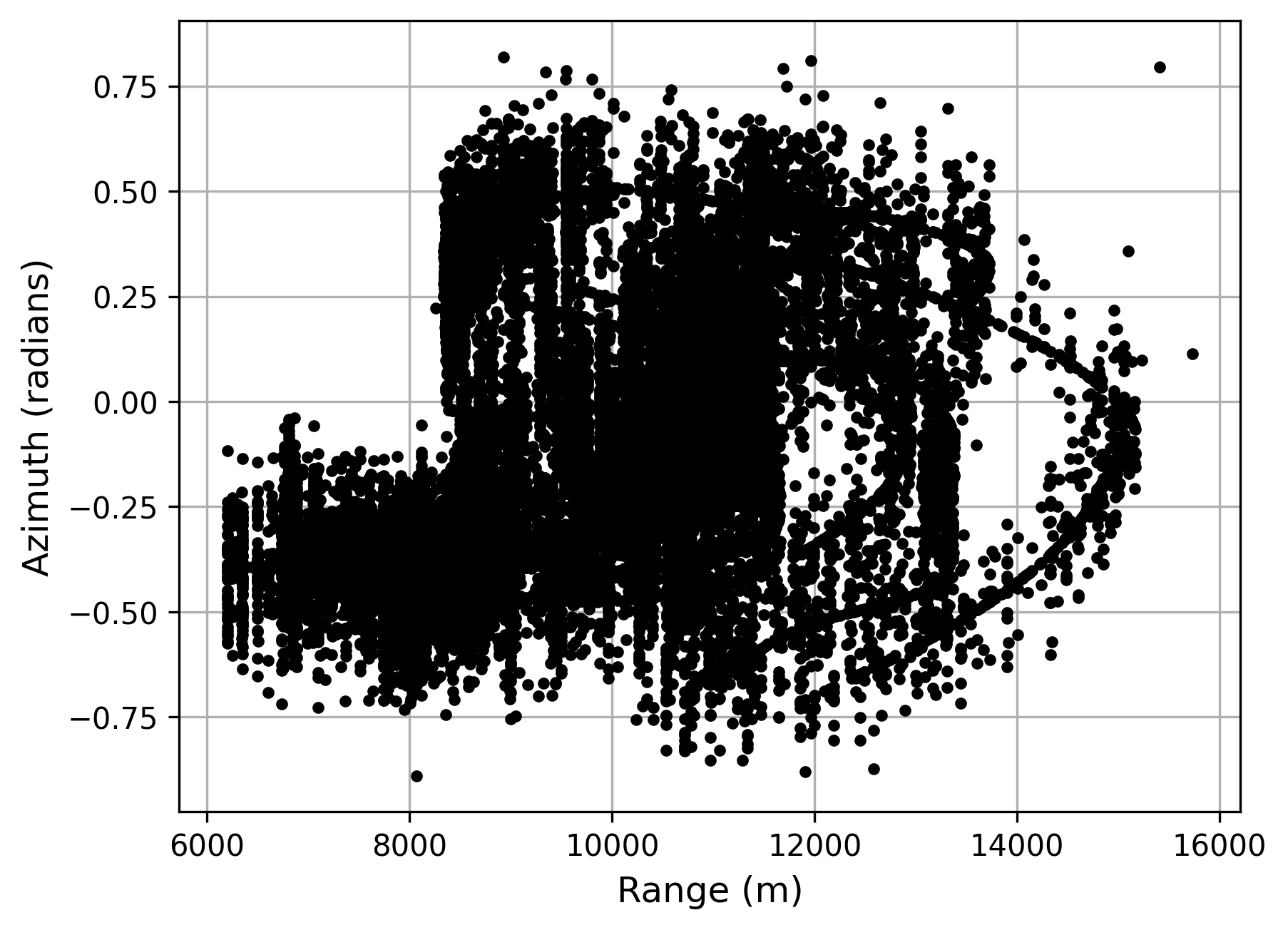}
    \subcaption{CFAR Test}
    \end{subfigure}
    \caption{Each black dot is a detection at some time point in the MoveStopMove scenario.}
    \label{fig:track_detections}
\end{figure}

Because our proposed machine learning methods have more accurate estimates of the uncertainty along with fewer false positive detections, it is able to give more accurate inputs to the tracker. An example of this is shown in Figure \ref{fig:track_detections} for the complex scenario with MoveStopMove dynamics where the CFAR test detects far more false positives, which are difficult for the filter to remove. 

\begin{figure}[H]
    \centering
    \begin{subfigure}{\linewidth}
        \centering
        \includegraphics[width=\linewidth]{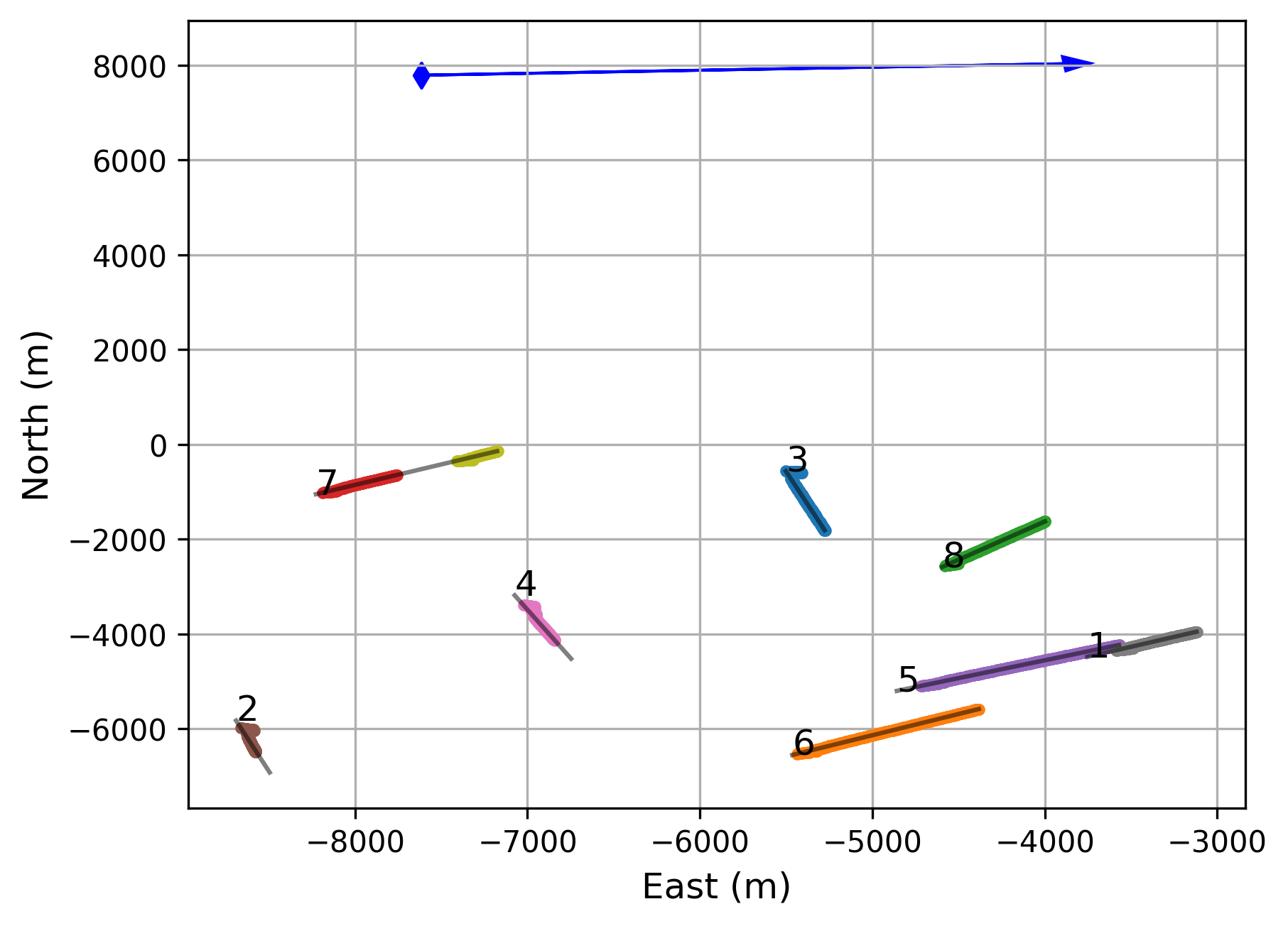}
        \caption{ML-Filter}
    \end{subfigure}
    \hfill
    \begin{subfigure}{\linewidth}
        \centering 
        \includegraphics[width=\linewidth]{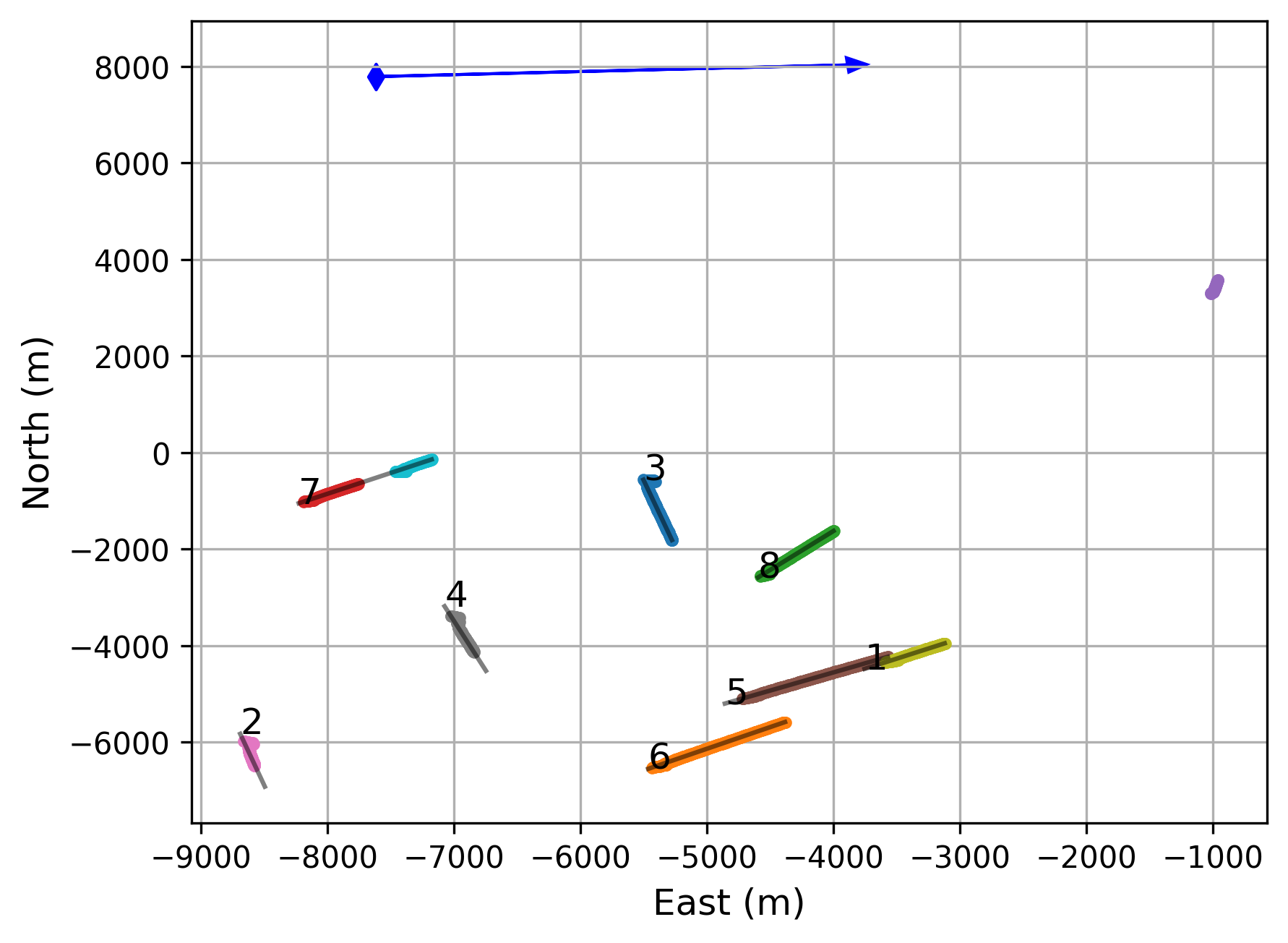}
        \caption{Baseline-Filter}
    \end{subfigure}
    \caption{Constant Velocity Dynamics (Simple Scenario)}
    \label{fig:simple_results}
\end{figure}

\begin{figure}[H]
    \begin{subfigure}{\linewidth} 
        \centering
        \includegraphics[width=\linewidth]{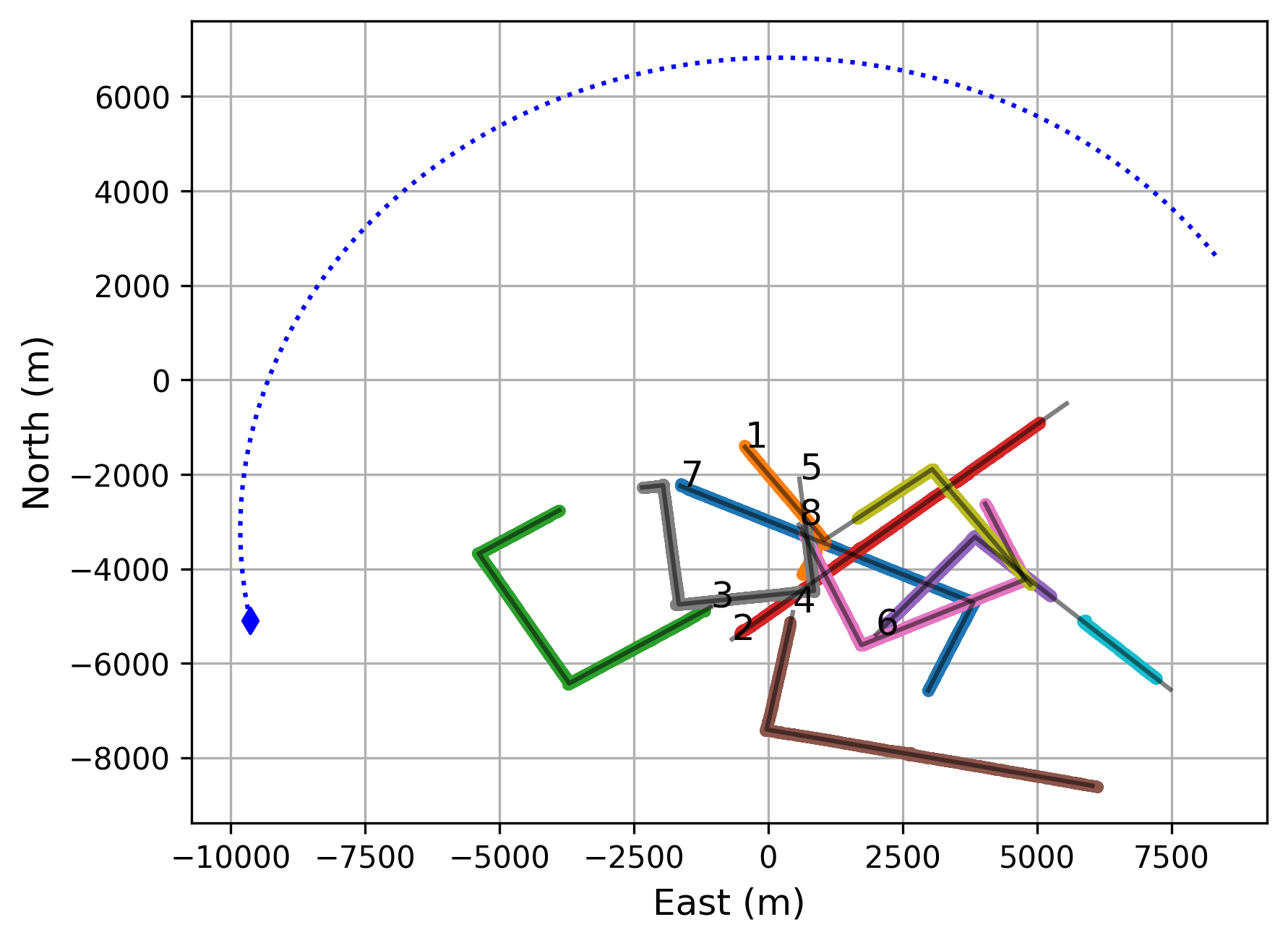}
        \caption{ML-Filter}
    \end{subfigure}
    \hfill
    \begin{subfigure}{\linewidth}  
        \centering 
        \includegraphics[width=\linewidth]{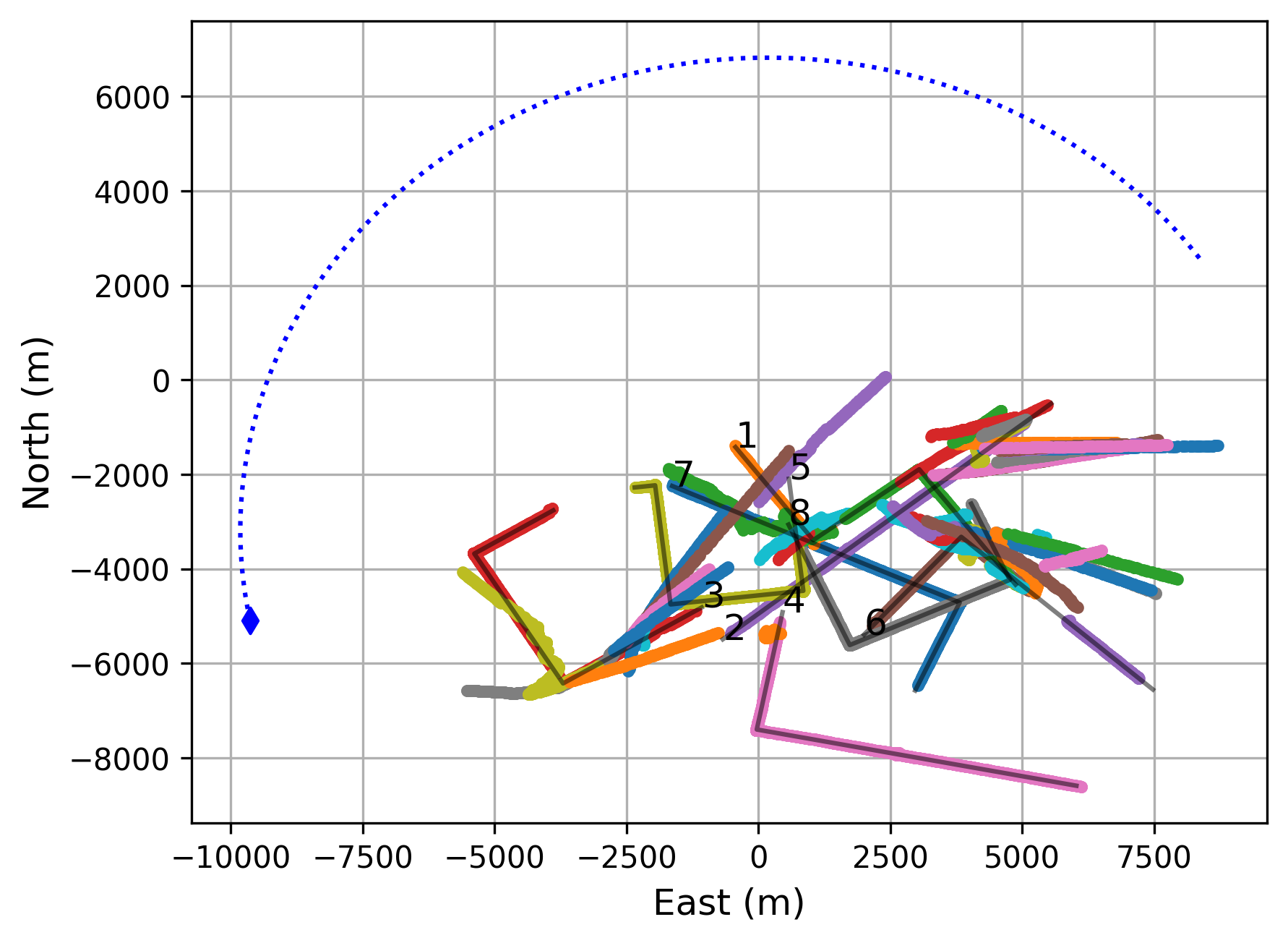}
        \caption{Baseline-Filter}
    \end{subfigure}
    \caption{MoveStopMove Dynamics (Complex Scenario)}
    \label{fig:complex_results}
\end{figure}

For visualization purposes, we also show the predicted tracks overlaid on the true target tracks for both methods for the simple scenario with constant velocity dynamics in Figure \ref{fig:simple_results} and the more complex scenario with MoveStopMove dynamics in Figure \ref{fig:complex_results}. In the figures, the blue diamond and arrow / dotted line represent the path of the platform over the scenario's time period, different colored points represent different unique predicted tracks, and the black dashed lines represent the true target tracks. The Baseline-Filter has visually much worse performance with an inclusion of an incorrect track (light purple) around the $(-1000, 3500)$ coordinates in the simple scenario and a significant number of false tracks in the more complex scenario. 

\section{Computational Costs} \label{computation}

In this section, we discuss some of the computational burdens of using machine learning models, which as we have shown above have superior accuracy, at the trade-off of potential computational complexity. Both the UNet and CVAE based neural network models are separately trained using a Nvidia Quadro RTX 6000 with 24GB of GPU memory. For each detection the UNet outputs a 32 dimensional feature vector, which is the primary large input (the other inputs are all scalars) for the CVAE model. However, this is for training the neural networks; for inference, we only need to consider the prediction capabilities of these networks, which can be done on CPU (we used a Ryzen Threadripper 3960X). Additionally, our experiments simulated sensors with only two channels where performing STAP is computationally cheap and relatively accurate. However, many real world radars have significantly larger numbers of channels, e.g. 20, making STAP both computationally much more expensive and significantly less accurate due to having to estimate high (e.g. 20) dimensional covariance matrices at every pixel in the endo-clutter region.

We summarize the computational costs in Table \ref{table:results} below where the inference time and FLOPs are calculated per input. 
\begin{table}[H]
\caption{Computational Costs}
\begin{center}
\begin{tabular}{l|llll}
                      & Memory & Disk Space & Inference Time & FLOPs \\
\hline
UNet         & 6.921 GB     & 99.0 MB         & 0.54 sec  & 761 GFLOPs \\
CVAE & 1.569 GB     & 164 KB & 0.095 sec  & 2.32 MFLOPS \\
\end{tabular}
\end{center}
\label{table:results}
\end{table}

\section{Conclusion} \label{end}
In this paper, we proposed two machine learning models i) a UNet based neural network with superior performance at detecting targets in RDM images, and ii) a CVAE based neural network for estimating the uncertainty around the target detector's predictions. Through experiments, we demonstrated that by using these machine learning models in the filtering step of a multiple hypothesis tracker, we are able to significantly improve the accuracy of the estimated tracks relative to using a theoretically (but not practically) optimal baseline system. As future work, we plan to further integrate our machine learning models into one unified model. Another interesting line of research is to combine the predictions and their uncertainties from multiple sensors (perhaps with different modalities) in a manner such as to improve the overall performance in downstream applications.

\appendix
\section{Appendix}

\subsection{The Measurement Model $H$} \label{h_func}

The dynamics model $\Phi$ generates a series of target locations and velocities in ENU for each target $k$ as $\{z^k_1, \dots, z^k_T\}$ where each $z^k_t$ is a 6 dimensional vector $[p_{e}, p_{n}, p_{u}, v_{e}, v_{n}, v_{u}]$. Let $h(\cdot)$ be a function that converts each $z^k_t$ to the measurement space that is being captured in RDM images, which has axes range and range-rate relative to a radar on a platform. So $h(\cdot)$ maps $z^k_t$ to a 4 dimensional measurement vector $y^k_t$ whose elements are
\begin{flalign}
& y_{range} = \big\lVert [\Delta p_{e}, \Delta p_{n}, \Delta p_{u}] \big\rVert_2 \label{range} \\
& y_{range-rate} = \frac{1}{y_{range}} [\Delta p_{e}, \Delta p_{n}, \Delta p_{u}] \begin{bmatrix} \Delta v_{e} \\ \Delta v_{n} \\ \Delta v_{u} \end{bmatrix} \label{rangerate} \\
& y_{azimuth} = \tan^{-1} \left(\frac{\Delta p_{e}}{\Delta p_{n}} \right) \label{azimuth} \\
& y_{elevation} = \sin^{-1} \left( \frac{\Delta p_{u}}{y_{range}} \right)
\end{flalign}
where $\Delta$ indicates relative to the platform position or velocity correspondingly. 

However, because $h(\cdot)$ is non-linear, we use its gradient evaluated at each target $k$ and time point $t$ for the measurement matrix $H$ in \eqref{measurement} (in a similar fashion as the extended Kalman filter). Explicitly
\begin{flalign}
H^k_t = \frac{\partial h}{\partial z} \bigg|_{z^k_t}
\end{flalign}
where $z^k_t$ is the ``predicted" state vector at time point $t$.

\subsection{Traditional Measurement Covariance Matrix}

For the baseline covariance matrix, we use the traditional method for estimating the measurement covariance as a function of the radar parameters. Namely,  
\begingroup
\addtolength{\arraycolsep}{-2pt} 
\begin{flalign}
\begin{bmatrix}
\frac{3}{4 \text{ SNR}} \frac{c^2}{4B^2}& 0 & 0 & 0\\
0 & \frac{3}{4 \text{ SNR}} \frac{\lambda^2}{4 (N_{pulse} T_{pri})^2} & 0 & 0 \\
0 & 0 & \frac{3}{4 \text{ SNR}} \frac{\lambda^2}{L_{az}^2} & 0 \\
0 & 0 & 0 & \frac{3}{4 \text{ SNR}} \frac{\lambda^2}{L_{el}^2} 
\end{bmatrix} \label{base_cov}
\end{flalign}
\endgroup
where SNR is the signal to noise ratio, $c$ is the speed of light, $B$ is the waveform bandwidth, $\lambda$ is the wavelength, $N_{pulse}$ is the number of pulses, $T_{pri}$ is the pulse repetition interval, and $L_{az}$ and $L_{el}$ are the dimensions of the physical radar array.

\end{document}